\documentclass[conference]{IEEEtran}
\IEEEoverridecommandlockouts
\usepackage{cite}
\usepackage{amsmath,amssymb}
\usepackage{booktabs}
\usepackage{multirow}
\usepackage{array}
\usepackage{graphicx}
\usepackage{xcolor}
\usepackage{url}
\usepackage{tikz}
\usetikzlibrary{positioning,arrows.meta,calc}
\usepackage[hidelinks]{hyperref}

\definecolor{cgblue}{HTML}{2A78D6}
\definecolor{cgorange}{HTML}{EB6834}
\definecolor{cgaqua}{HTML}{1BAF7A}
\definecolor{cggray}{HTML}{52514E}

\newcommand{\sys}{\textsc{ClaimGate}}
\graphicspath{{figures/}}

\begin{document}

\title{\sys: Cautious On-Device Checking of Spoken Legal and Clinical Claims}

\author{\IEEEauthorblockN{Peter Forcha}
\IEEEauthorblockA{\textit{Dept. of Electrical and Computer Engineering}\\
\textit{University of Florida}\\
Gainesville, FL, USA\\
peter.forcha@ufl.edu}
\and
\IEEEauthorblockN{Harshitha Kaje Kusumadhara}
\IEEEauthorblockA{\textit{Dept. of Electrical and Computer Engineering}\\
\textit{University of Florida}\\
Gainesville, FL, USA\\
h.kajekusumadhar@ufl.edu}
\and
\IEEEauthorblockN{Audrey Moguem Souop}
\IEEEauthorblockA{\textit{Dept. of Electrical and Computer Engineering}\\
\textit{University of Florida}\\
Gainesville, FL, USA\\
audreycyriell.mo@ufl.edu}
\and
\IEEEauthorblockN{Shalom Iboh}
\IEEEauthorblockA{\textit{Dept. of Electrical and Computer Engineering}\\
\textit{University of Florida}\\
Gainesville, FL, USA\\
shalom.iboh@ufl.edu}}

\maketitle

\begin{abstract}
In law and medicine, a misstated rule can shape a decision before anyone has time to check it. Checking such claims in real time helps only if the checker keeps pace with conversation, keeps sensitive audio on the device, and stays silent rather than mislead. We present \sys{}, an on-device assistant that listens to a conversation, detects checkable claims, retrieves the relevant official rule, and labels each claim as supported, contradicted or not covered, with a citation. \sys{} is designed to fail towards silence: its 4-bit 7B-parameter language model may cite only retrieved rules, and the verdict is derived in code from the claim and rule the model quotes. We evaluate it on traffic stops and medical visits using real official rule text (5,742 Florida and Texas statute subsections; CDC, FDA and MedlinePlus guidance), real transcripts, live speech and author-written gold-labelled claims. On 141 held-out traffic-stop claims, \sys{} reaches 80.1\% accuracy and cuts serious errors (a true claim called false, or the reverse) from 20 to 5 compared with the same model without retrieval; deriving the verdict in code alone reduces them from 13 to 5 at no cost to accuracy. The claim gate passes 95.7\% of held-out claims and rejects all 30 everyday sentences, and feedback arrives a median 1.86\,s after a speech window closes on an 8\,GB consumer GPU. On 148 allergy claims, with only the claim gate to rebuild, \sys{} reaches 87.8\% accuracy with 5 serious errors, against 20 without retrieval.
\end{abstract}


\section{Introduction}
\label{sec:intro}

People often make statements about laws and medical guidelines in conversation, and these statements are not always accurate. During a traffic stop, a driver might misstate when police can search a vehicle; in a clinic, a patient might misremember guidance on allergies and vaccination. Verifying such statements means searching lengthy statutes or guidelines, which nobody can do mid-conversation. Identifying and checking them in real time could help drivers, officers, patients and clinicians make better-informed decisions.

Building such a system poses three challenges. First, most speech contains no checkable claim, so claim detection must be accurate. Second, a wrong verdict can have serious consequences, so the system must avoid serious errors and abstain rather than issue a verdict it cannot ground in a rule. Third, these conversations are sensitive, so processing must run privately on local hardware, within the memory and latency budget of a single consumer GPU or embedded board.

We introduce \sys{}, an on-device system that converts speech into text, identifies checkable claims, retrieves relevant legal or medical text, and uses a local language model to label each claim supported, contradicted or not covered, with a citation. Its verifier uses a constrained restate--quote--match design: the model restates the claim, names one retrieved rule, quotes it, and judges whether the two match; the verdict is then derived from that judgment in code, and any verdict that does not cite a retrieved rule is downgraded to \emph{not covered}. When no relevant rule is retrieved, or the model cites none, the system abstains.

We evaluate \sys{} on traffic-stop claims, against official 2026 Florida and Texas statutes and federal case law, and on allergy claims, against CDC, FDA and MedlinePlus guidance. Since labelled recordings of real stops and consultations are not available, labelled claims are written by the authors in a spoken style; real meeting transcripts, human-written dialogue and live microphone sessions test claim detection and end-to-end behavior. On 141 held-out legal claims, \sys{} achieves 80.1\% accuracy with five serious errors, against 20 for the same model without retrieval, and 87.8\% on 148 allergy claims without retuning the verifier.

Our contributions are:
\begin{itemize}
\item An end-to-end on-device pipeline that detects and verifies spoken claims against authoritative documents within an 8\,GB GPU budget, with a median latency of 1.86\,s per claim.
\item A cautious verification strategy combining retrieval, grammar-constrained model output, programmatic verdicts and abstention.
\item An evaluation of claim detection, retrieval, verification, latency and memory using real source documents, real conversational speech and carefully constructed synthetic claims.
\item A two-domain evaluation showing that the verifier transfers and that the claim gate is the one component that must be rebuilt per domain.
\end{itemize}

\section{Background and Related Work}
\label{sec:related}

\textit{Automated fact-checking.}
Fact-checking commonly involves identifying check-worthy claims, retrieving evidence, and assessing claims against it~\cite{guo2022survey,nakov2021survey}.
ClaimBuster~\cite{hassan2017claimbuster} identifies claims worth checking in political debates, and FEVER~\cite{thorne2018fever} introduced a widely used framework with three labels: supported, refuted and not enough information, which we adapt as \emph{supported}, \emph{contradicted} and \emph{not covered}.
Unlike general-purpose systems, \sys{} checks claims made in live speech against a bounded collection of authoritative legal and clinical sources, and treats a wrong verdict as costlier than none.

\textit{Legal NLP and hallucination.}
LexGLUE~\cite{chalkidis2022lexglue} and LegalBench~\cite{guha2023legalbench} benchmark legal language understanding and reasoning, but benchmark performance does not guarantee reliable answers in practice.
Dahl et al.~\cite{dahl2024fictions} document frequent legal hallucinations in general-purpose models, and Magesh et al.~\cite{magesh2024hallucination} show that retrieval-based legal research tools also produce unsupported information.
Our closed-book baseline reproduces this: without retrieved rules, the same model makes 20 serious errors on our 141 held-out claims, against 5 for the full system (Table~\ref{tab:main}).

\textit{Retrieval augmentation and constrained decoding.}
Retrieval-augmented generation (RAG) gives a model relevant passages before it answers~\cite{lewis2020rag}.
\sys{} uses dense retrieval with bge-small-en-v1.5~\cite{xiao2023cpack}, compared against BM25~\cite{robertson2009bm25} and a hybrid.
The verifier's output follows a JSON schema compiled into a grammar~\cite{willard2023guided,gerganov2023llamacpp}, which restricts rule selection to the retrieved passages; code maps the model's comparison to the final label, and if no rule is selected the system returns \emph{not covered}.

\textit{Selective prediction.}
Letting a model abstain trades coverage for lower error risk~\cite{geifman2017selective}; Kamath et al.~\cite{kamath2020selective} use a learned calibrator to abstain in question answering under domain shift.
\sys{} instead builds abstention into the task: \emph{not covered} is always an allowed answer and is the fallback when no rules are retrieved, no rule is cited, or the output cannot be parsed. Apart from a retrieval relevance threshold, no calibrated confidence model is used.

\section{Problem Formulation}
\label{sec:problem}

The input is a stream of speech and the jurisdiction in which it happens.
The system must (i) find windows that contain a checkable claim, and (ii) for each, return \emph{supported}, \emph{contradicted} or \emph{not covered}, with the one rule the verdict rests on and a one-line explanation.
\emph{Not covered} means the rules found do not settle the claim, not that it is false.
A \textbf{serious error} is a supported claim labelled contradicted or a contradicted claim labelled supported.
The design objectives, in priority order, are: few serious errors; no verdicts on speech that contains no claim; a verifiable citation for every verdict; high claim recall; and feedback within seconds on local hardware, with no audio leaving the device.

\section{The \sys{} Framework}
\label{sec:system}

\begin{figure*}[t]
\centering
\resizebox{0.86\textwidth}{!}{%
\begin{tikzpicture}[
  font=\small,
  stage/.style={draw=cgblue, fill=cgblue!8, rounded corners=3pt, align=center, minimum height=1.25cm, text width=2.45cm, inner sep=3pt},
  io/.style={draw=cggray, fill=gray!8, rounded corners=3pt, align=center, minimum height=1.25cm, text width=2.0cm, inner sep=3pt},
  silent/.style={draw=cgorange, fill=cgorange!8, rounded corners=3pt, align=center, text width=2.3cm, inner sep=3pt, font=\footnotesize},
  arr/.style={-{Stealth[length=5pt]}, thick, draw=cggray},
  drop/.style={-{Stealth[length=4pt]}, draw=cgorange, dashed},
  node distance=0.5cm]
  \node[io] (mic) {\textbf{Speech}\\[1pt]\footnotesize driver, officer,\\patient, clinician};
  \node[stage, right=of mic] (asr) {\textbf{Streaming ASR}\\[1pt]\footnotesize whisper.cpp base.en\\(CPU)};
  \node[stage, right=of asr] (win) {\textbf{Windowing}\\[1pt]\footnotesize 3.5\,s pause\\or 35 words};
  \node[stage, right=of win] (gate) {\textbf{Claim gate}\\[1pt]\footnotesize TF-IDF+LR $\vee$\\BERT-tiny, 2.9\,ms};
  \node[stage, below=1.6cm of gate] (ret) {\textbf{Dense retrieval}\\[1pt]\footnotesize bge-small, state +\\federal, top-5 sections};
  \node[stage, left=of ret] (rel) {\textbf{Relevance check}\\[1pt]\footnotesize best cosine\\$\ge 0.544$};
  \node[stage, left=of rel] (ver) {\textbf{Verifier LLM}\\[1pt]\footnotesize Qwen2.5-7B Q4,\\JSON grammar};
  \node[stage, left=of ver, fill=cgaqua!10, draw=cgaqua] (map) {\textbf{Verdict in code}\\[1pt]\footnotesize same/different/\\unrelated $\to$ label};
  \node[io, right=0.6cm of ret, text width=2.3cm] (corp) {\textbf{Real rule text}\\[1pt]\footnotesize 5,742 statute\\subsections + 22\\case holdings};
  \node[draw=cgaqua, fill=cgaqua!10, rounded corners=3pt, align=left, below=0.7cm of $(map.south)!0.5!(rel.south)$, text width=12cm, font=\footnotesize, inner sep=5pt] (out)
   {\textbf{Example (held-out claim, real output):}\ ``you only need to signal 20 feet before a turn'' $\to$ \emph{Contradicted} --- Fla.\ Stat.\ \S 316.155: ``A signal of intention to turn right or left must be given continuously during not less than the last 100 feet traveled by the vehicle before turning.''};
  \draw[arr] (mic) -- (asr);
  \draw[arr] (asr) -- (win);
  \draw[arr] (win) -- (gate);
  \draw[arr] (gate) -- node[right, font=\footnotesize]{claim} (ret);
  \draw[arr] (corp) -- (ret);
  \draw[arr] (ret) -- (rel);
  \draw[arr] (rel) -- (ver);
  \draw[arr] (ver) -- (map);
  \draw[arr] (map.south) |- (out.west);
  \node[silent, above=0.32cm of gate] (s1) {no claim $\to$ silent};
  \draw[drop] (gate) -- (s1);
  \node[silent, above=0.32cm of rel] (s2) {weak match $\to$ silent};
  \draw[drop] (rel) -- (s2);
  \node[silent, above=0.32cm of ver] (s3) {no rule cited $\to$\\\emph{not covered}};
  \draw[drop] (ver) -- (s3);
\end{tikzpicture}%
}
\caption{The \sys{} pipeline. Blue stages run on every window; dashed orange arrows are the exits that make the system fall silent or abstain instead of guessing. Everything runs on one desktop with an 8\,GB consumer GPU; no audio leaves the device.}
\label{fig:pipeline}
\end{figure*}
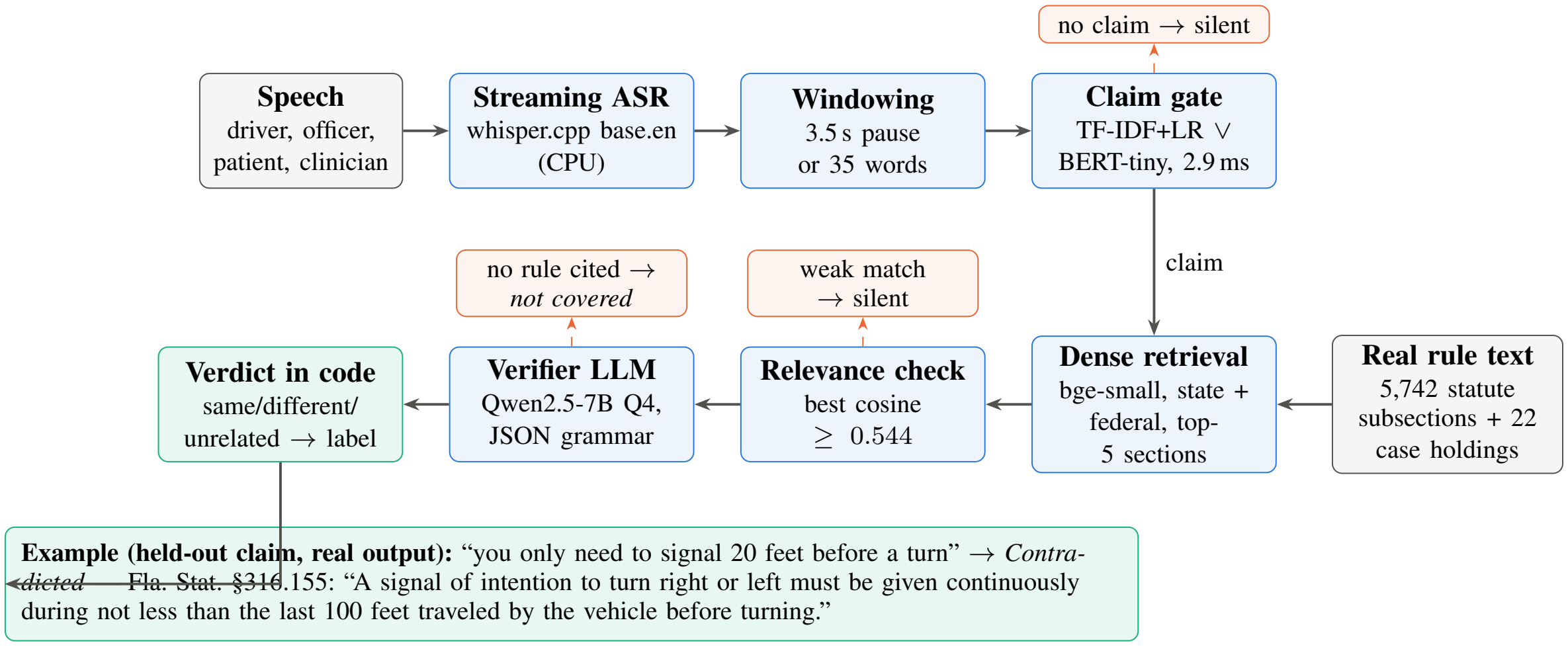

Fig.~\ref{fig:pipeline} shows the pipeline. Each stage is a cheap filter in front of a more expensive one: the claim gate (milliseconds) and relevance check (microseconds) decide whether the verifier LLM (about two seconds) runs at all, so most speech never reaches the GPU.

\subsection{Speech and Windowing}
whisper.cpp runs Whisper base.en~\cite{radford2023whisper} on the CPU in streaming mode, and hypotheses are merged into word windows that close after a 3.5\,s pause or at 35 words.
Raw Whisper text goes to the gate; we do not clean transcripts with an LLM, because a rewriting model can change a claim before it is checked.

\subsection{Claim Gate}
The gate is the logical OR of TF-IDF with logistic regression and a fine-tuned BERT-tiny~\cite{devlin2019bert,turc2019wellread}.
Each threshold is set for 95\% recall on a validation split built from templates that training never uses, which raised test recall by about ten points over a random split.
With only traffic-stop chatter as negatives, the gate fired on 77\% of windows of a real, unrelated lab-meeting transcript; the deployed gate adds 400 template off-topic statements and 1,000 everyday-conversation windows from DailyDialog~\cite{li2017dailydialog} as negatives.

\subsection{Rule Corpus, Retrieval and Relevance Check}
The corpus is the official 2026 text of Florida Statutes chapters 316, 318, 320, 322 and 324 (plus three sections) and Texas Transportation Code chapters 502, 521, 542--548, 550, 551, 601 and 724 and Penal Code \S\S 16.02, 38.02 and chapter 49: 5,742 subsection-level records, about 505k words.
We add 22 hand-written summaries of US Supreme Court and circuit holdings that govern stops (e.g.\ \emph{Whren}, \emph{Rodriguez}, \emph{Mimms}, \emph{Gant}).
Records are split into 75-word chunks with 15 words of overlap, each prefixed with its section heading, and embedded with bge-small-en-v1.5.
A query searches only the stop's state plus federal law and returns the top five \emph{distinct} sections; on development data, headings were worth about 9 points of top-5 recall and distinct sections 4--5 more.
A window goes to the verifier only if its best chunk has cosine similarity $\geq 0.544$, the lowest value of any development claim.

\subsection{Constrained Verifier and Programmatic Verdict}
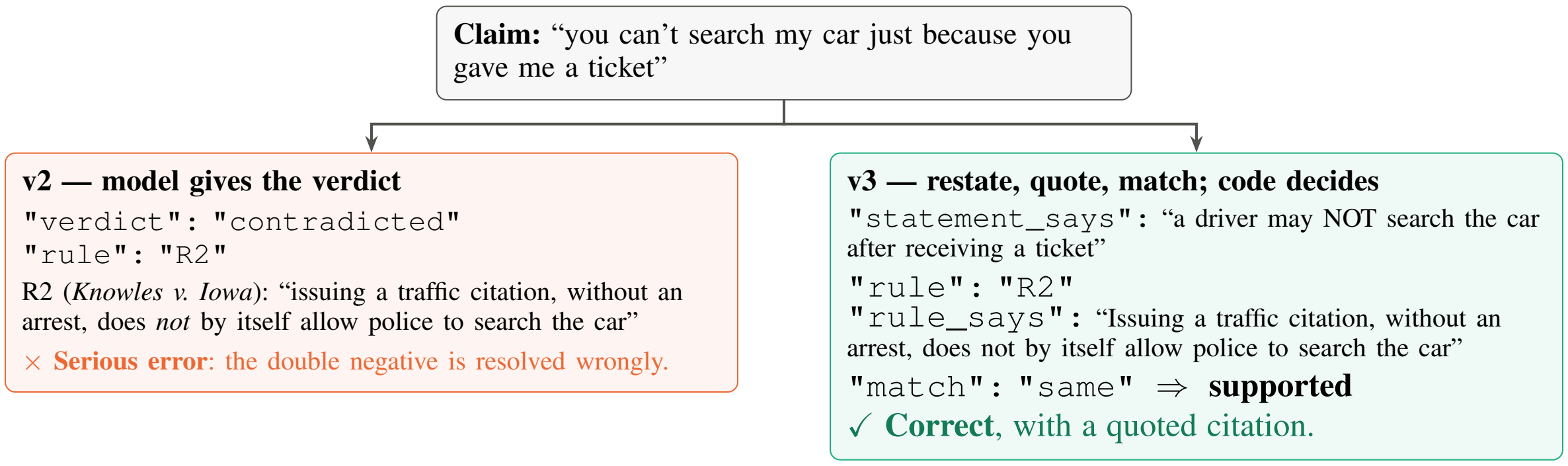
\begin{figure*}[t]
\centering
\resizebox{0.86\textwidth}{!}{%
\begin{tikzpicture}[
  font=\small,
  box/.style={draw, rounded corners=3pt, align=left, inner sep=5pt},
  arr/.style={-{Stealth[length=5pt]}, thick, draw=cggray}]
  \node[box, draw=cggray, fill=gray!6, text width=7cm] (claim)
    {\textbf{Claim:} ``you can't search my car just because you gave me a ticket''};
  \node[box, draw=cgorange, fill=cgorange!7, text width=7.4cm, below left=0.55cm and -3.2cm of claim, anchor=north east] (v2)
    {\textbf{v2 --- model gives the verdict}\\[2pt]
     \texttt{"verdict": "contradicted"}\\
     \texttt{"rule": "R2"}\\[2pt]
     \footnotesize R2 (\emph{Knowles v.\ Iowa}): ``issuing a traffic citation, without an arrest, does \emph{not} by itself allow police to search the car''\\[2pt]
     \textcolor{cgorange}{\textbf{$\times$ Serious error}: the double negative is resolved wrongly.}};
  \node[box, draw=cgaqua, fill=cgaqua!7, text width=7.4cm, below right=0.55cm and -3.2cm of claim, anchor=north west] (v3)
    {\textbf{v3 --- restate, quote, match; code decides}\\[2pt]
     \texttt{"statement\_says":} \footnotesize``a driver may NOT search the car after receiving a ticket''\\
     \normalsize\texttt{"rule": "R2"}\\
     \texttt{"rule\_says":} \footnotesize``Issuing a traffic citation, without an arrest, does not by itself allow police to search the car''\\
     \normalsize\texttt{"match": "same"}\ \ $\Rightarrow$\ \ \textbf{supported}\\[2pt]
     \textcolor{cgaqua!70!black}{\textbf{\checkmark\ Correct}, with a quoted citation.}};
  \draw[arr] (claim.south) -- ++(0,-0.25) -| (v2.north);
  \draw[arr] (claim.south) -- ++(0,-0.25) -| (v3.north);
\end{tikzpicture}%
}
\caption{The two verifier prompts on a negated held-out claim (outputs abridged from the actual runs). v3 makes the model rewrite the claim as a plain rule and quote the rule before stating whether they match; the label is mapped in code. \texttt{rule} is a grammar-enforced enum over the shown rules, so invented citations are impossible.}
\label{fig:prompt}
\end{figure*}

Qwen2.5-7B-Instruct~\cite{qwen2024qwen25}, quantized to 4 bits (Q4\_K\_M), runs in llama.cpp~\cite{gerganov2023llamacpp} on the GPU.
Each of the five rules is shown as its retrieved subsection followed by the rest of its section, cut to 220 words, so that a definition is seen together with the duty it qualifies.
A JSON schema forces four fields, filled in order (Fig.~\ref{fig:prompt}):
\texttt{statement\_says} (the claim restated as a plain rule, keeping every \emph{not}, \emph{only} and number),
\texttt{rule} (an enum of \texttt{R1}--\texttt{R5} or \texttt{none}),
\texttt{rule\_says} (the deciding words of that rule, quoted) and
\texttt{match} (\emph{same}, \emph{different} or \emph{unrelated}).
Code maps \emph{same}$\to$supported, \emph{different}$\to$contradicted and \emph{unrelated}$\to$not covered; a verdict with no rule becomes not covered.
Because \texttt{rule} is an enum, the model cannot cite a rule it was not given.
We call this prompt \textbf{v3}; the earlier \textbf{v2} asks for the verdict directly.

\section{Experimental Setup}
\label{sec:setup}

\begin{table}[t]
\centering
\caption{Evaluation data: a mixture of real and synthetic sources.}
\label{tab:data}
\footnotesize
\setlength{\tabcolsep}{3pt}
\begin{tabular}{@{}>{\raggedright}p{2.45cm}>{\raggedright}p{3.55cm}c@{}}
\toprule
Use & Data & Type \\
\midrule
\multicolumn{3}{@{}l}{\emph{Evidence (what the system checks against)}} \\
Legal rules & 5,742 FL/TX statute subsections (official 2026 text) & Real \\
Case law & 22 summaries of federal holdings & Author-written \\
Clinical rules & 280 passages: CDC, MedlinePlus, FDA, 17 openFDA labels & Real \\
\midrule
\multicolumn{3}{@{}l}{\emph{Speech and conversation}} \\
Off-topic test & 2 real meeting transcripts (66 / 183 windows) & Real \\
Gate negatives & 1,000 DailyDialog windows & Real (human-written) \\
End-to-end pilot & 137 live microphone windows, 7 sessions & Real \\
\midrule
\multicolumn{3}{@{}l}{\emph{Labelled claims}} \\
Gate train / val & 1,800 / 200 template windows & Synthetic \\
Gate test & 1,001 windows (525 claims); 60-window challenge set & Synthetic \\
Retrieval & 67 dev / 50 held-out claims & Synthetic \\
Verification (legal) & 46 dev / 141 held-out claims; 50 non-claims & Synthetic \\
Verification (allergy) & 50 dev / 148 test claims; 50 non-claims & Synthetic \\
\bottomrule
\end{tabular}
\end{table}

\subsection{Hardware Platform}
All experiments run on one desktop with an Intel i7-9700K CPU and an NVIDIA RTX 2070 GPU (8\,GB). Whisper, the gate and the retriever run on the CPU and the verifier on the GPU; no stage calls a network service.

\subsection{Datasets}
Table~\ref{tab:data} lists every data source and whether it is real or synthetic.
All rule text is real and official.
Two real lab-meeting transcripts with no traffic claims serve as development and test sets for off-topic speech, and 1,000 human-written DailyDialog windows serve as gate negatives.
We also ran the full pipeline on live microphone speech in 7 sessions (137 windows; 48 passed the gate and 42 received a verdict); these are unlabelled and used qualitatively (Section~\ref{sec:errors}).

Labelled claims are synthetic: written by the authors in a spoken style and checked against the corpus, since labelled recordings of real stops are not available.
The gate test set (525 claims) was converted from a separately written collection of traffic-stop assertions.
Verification uses 187 claims with a gold verdict and gold section: dev (26) and dev2 (20) for development, test (24) and test2 (117) held out.
The 141 held-out claims are 66 supported, 59 contradicted and 16 not covered.
test2 was written before the tuning reported here and frozen with a SHA-256 hash, and has 50 added non-claim lines: 30 everyday sentences and 20 stop-talk lines with no legal claim (``license and registration please'').

\subsection{Protocol and Metrics}
All choices were made on development data using a rule fixed in advance: fewest serious errors, then accuracy.
We report 95\% percentile bootstrap intervals (10,000 resamples), paired bootstrap intervals for differences~\cite{koehn2004significance,dror2018hitchhiker} and Wilson intervals for rates~\cite{wilson1927}.
Besides accuracy and serious errors we report \emph{coverage} (share given supported/contradicted), \emph{risk} (serious errors among those), contradiction recall, citation accuracy, latency and memory.

\section{Experimental Results}
\label{sec:results}

\begin{table*}[t]
\centering
\caption{Held-out verification (test+test2, 141 claims, real retrieval unless noted). Serious: supported$\leftrightarrow$contradicted errors. $\Delta$: paired difference to the full system; * marks a 95\% paired bootstrap interval that excludes zero. Cov.: share of claims given supported/contradicted. Risk: serious errors among those. C-rec: recall of contradicted claims. Cite: correct citation among correct supported/contradicted verdicts.}
\label{tab:main}
\footnotesize
\begin{tabular}{lcccccccc}
\toprule
System & Accuracy & Serious & $\Delta$Acc & $\Delta$Serious & Cov. & Risk & C-rec & Cite \\
\midrule
\textbf{Full system} (v3, dense, $k{=}5$, section) & \textbf{0.801} {\scriptsize[.738,.865]} & \textbf{5} {\scriptsize[1,9]} & --- & --- & 0.72 & 0.049 & 0.66 & 0.85 \\
\midrule
\multicolumn{9}{l}{\emph{Baselines}} \\
No retrieval (closed book, same model) & 0.596 {\scriptsize[.511,.674]} & 20 {\scriptsize[12,29]} & $-$0.206* & +15* & 0.68 & 0.208 & 0.75 & --- \\
Prompt v2 (model gives the verdict) & 0.787 {\scriptsize[.716,.851]} & 13 {\scriptsize[7,20]} & $-$0.014 & +8* & 0.77 & 0.120 & 0.76 & 0.75 \\
BM25 retrieval & 0.660 {\scriptsize[.582,.738]} & 9 {\scriptsize[4,15]} & $-$0.142* & +4 & 0.62 & 0.102 & 0.49 & 0.83 \\
Hybrid retrieval (dense + BM25) & 0.780 {\scriptsize[.709,.844]} & 10 {\scriptsize[4,16]} & $-$0.021 & +5 & 0.74 & 0.096 & 0.66 & 0.86 \\
\midrule
\multicolumn{9}{l}{\emph{Ablations}} \\
Top-1 rule & 0.745 {\scriptsize[.674,.816]} & 13 {\scriptsize[7,20]} & $-$0.057 & +8* & 0.74 & 0.125 & 0.69 & 0.74 \\
Top-3 rules & 0.738 {\scriptsize[.660,.809]} & 14 {\scriptsize[7,21]} & $-$0.064 & +9* & 0.72 & 0.137 & 0.64 & 0.86 \\
Top-10 rules & 0.851 {\scriptsize[.787,.908]} & 10 {\scriptsize[5,16]} & +0.050 & +5 & 0.82 & 0.086 & 0.78 & 0.84 \\
Chunk context (75 words, no section) & 0.759 {\scriptsize[.688,.830]} & 15 {\scriptsize[8,23]} & $-$0.043 & +10* & 0.78 & 0.136 & 0.68 & 0.78 \\
\midrule
\multicolumn{9}{l}{\emph{References}} \\
Oracle retrieval (gold rule forced in) & 0.858 {\scriptsize[.801,.915]} & 7 {\scriptsize[2,12]} & +0.057 & +2 & 0.79 & 0.062 & 0.75 & 0.89 \\
Same model, 3-bit (Q3\_K\_M, 3.6\,GB) & 0.759 {\scriptsize[.688,.830]} & 13 {\scriptsize[7,20]} & $-$0.043 & +8* & 0.74 & 0.125 & 0.61 & 0.87 \\
Qwen2.5-14B (Q4\_K\_M) & 0.830 {\scriptsize[.766,.894]} & 13 {\scriptsize[7,20]} & +0.028 & +8 & 0.82 & 0.112 & 0.88 & 0.87 \\
\bottomrule
\end{tabular}
\end{table*}

\begin{figure}[t]
\centering
\includegraphics[width=0.74\columnwidth]{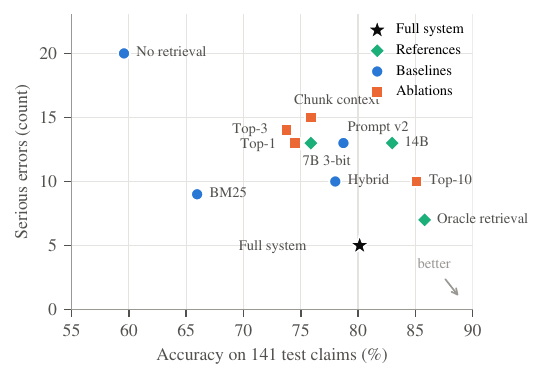}
\caption{Accuracy against serious errors for every system in Table~\ref{tab:main}. The full system has the fewest serious errors; more accurate configurations (top-10, oracle, 14B) commit to more verdicts and make more serious errors.}
\label{fig:tradeoff}
\end{figure}

\subsection{Verification}
Table~\ref{tab:main} and Fig.~\ref{fig:tradeoff} give the held-out results.
The full system is correct on 80.1\% of claims and makes 5 serious errors (3.5\% of claims; 4.9\% of the verdicts it commits to).
All 16 \emph{not covered} claims are answered correctly; most remaining errors are abstentions (16 contradicted and 7 supported claims answered \emph{not covered}), 13 of which follow a retrieval miss.

\textit{Retrieval is essential.}
Without retrieval the same model makes four times as many serious errors (20 vs.\ 5, paired difference $+15$, CI $[+7,+24]$) and is 20.6 points less accurate: its own knowledge of state traffic law is confident and often wrong~\cite{dahl2024fictions}.
BM25 costs 14 points of accuracy; the hybrid is not significantly different from dense retrieval.

\begin{figure}[t]
\centering
\includegraphics[width=0.74\columnwidth]{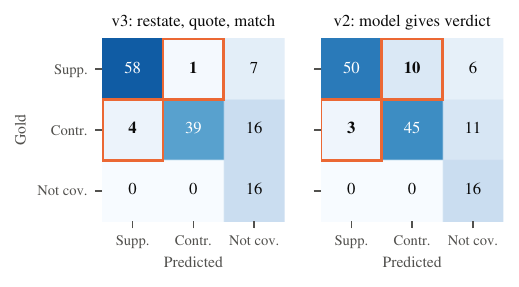}
\caption{Confusion matrices of the two prompts on the 141 held-out claims. Orange cells are serious errors (5 for v3, 13 for v2).}
\label{fig:confusion}
\end{figure}

\textit{Restate, quote, then match.}
Asking for the verdict directly (v2) gives the same accuracy but 13 serious errors instead of 5 ($+8$, $[+1,+15]$; Fig.~\ref{fig:confusion}).
Ten of v2's 13 serious errors call a true claim false, mostly negated claims (\emph{can't}, \emph{don't need}, \emph{only}), as in Fig.~\ref{fig:prompt}.
With v3 the comparison becomes a match between two sentences rather than the resolution of a double negative; v3 answers 11 of v2's 13 serious errors correctly and raises citation accuracy from 0.75 to 0.85.
It is more cautious: coverage 0.72 vs.\ 0.77, and 39 vs.\ 45 of 59 contradictions caught.
A second ``confirm'' call before contradicted verdicts discarded most correct contradictions on development data and was not adopted.

\subsection{Ablations}
Showing only the 75-word chunk instead of the section triples serious errors ($+10$, $[+5,+16]$).
Top-1 and top-3 add 8 and 9 serious errors, mainly because the deciding rule is more often missing; ten rules raise accuracy to 0.851 but double serious errors and add about a second per claim.
Forcing the gold rule in (oracle retrieval) raises accuracy by 5.7 points without reducing serious errors, so the remaining serious errors are mostly verifier errors.
At temperature zero with a fixed seed, four fresh-server runs gave identical verdicts on all 141 claims; one run on a long-running server differed on 5 claims, likely due to reused prompt-cache state, and the majority vote over all five runs equals the reported result.

\subsection{Claim Detection and Off-Topic Speech}

\begin{figure}[t]
\centering
\includegraphics[width=0.66\columnwidth]{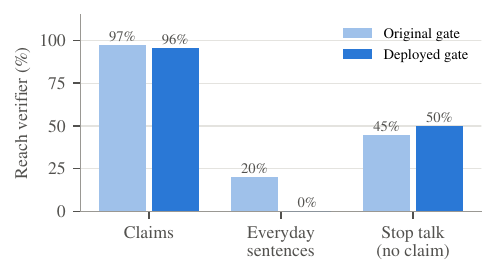}
\caption{Share of test2 lines that reach the verifier after the gate and relevance check. Off-topic negatives stop every everyday sentence at a cost of 1.7 points of claim recall.}
\label{fig:gate}
\end{figure}

With the relevance check, the deployed gate passes 112 of 117 test2 claims (0.957, CI 0.904--0.982) and none of 30 everyday sentences (CI 0--0.114), against 6 of 30 for the original gate (Fig.~\ref{fig:gate}).
On the real meeting transcript it lets 8.7\% of windows through, against 77\% for the original gate; on the earlier synthetic test sets claim recall is 0.962 (525 claims) and 0.900 (challenge set).
Four of the five missed claims concern police powers, which the gate's templates cover thinly.
Sent straight to the verifier, 6 of the 50 non-claim lines get a verdict, e.g.\ ``license and registration please'' \emph{supported} by a registration statute; behind the gate and relevance check, 3 do (6\%), all stop talk, and everyday sentences never reach the verifier.

\subsection{Retrieval}
On 67 development claims, dense retrieval finds a gold section in the top five for 92.5\% (top-1 61.2\%) in 9\,ms, against 86.6\% for a dense+BM25 hybrid and 50.7\% for BM25; without the state filter it falls to 79.1\%.
A cross-encoder reranker lowers top-1 to 37.3\%, and larger embedders did not help.
On 50 held-out retrieval claims top-5 recall is 88\%, and for the 141 held-out verification claims a gold rule is retrieved for 105 of the 125 that have one (84\%).

\begin{figure*}[t]
\centering
\includegraphics[width=0.6\textwidth]{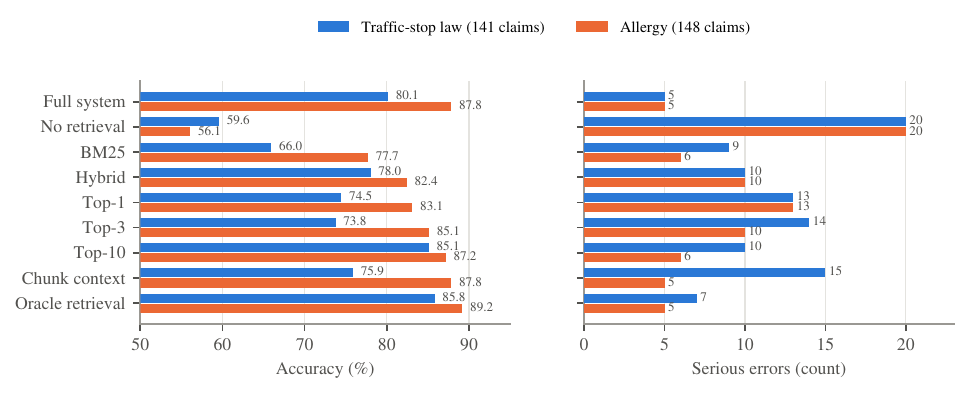}
\caption{The same systems in both domains, with no retuning for allergy. No retrieval is worst in both and the full system has the fewest serious errors in both. Chunk context hurts in law, where statute subsections depend on definitions elsewhere in the section, but not for self-contained guideline paragraphs.}
\label{fig:domains}
\end{figure*}

\subsection{Latency and Memory}
\label{sec:latency}

\begin{figure}[t]
\centering
\includegraphics[width=0.62\columnwidth]{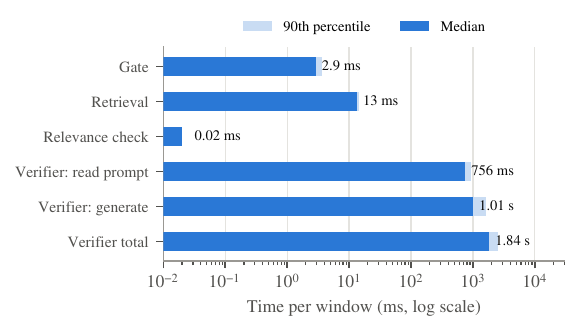}
\caption{Per-window time of each stage (log scale; i7-9700K + RTX 2070). The verifier accounts for over 99\% of the time to feedback.}
\label{fig:latency}
\end{figure}

Median per-window times are 2.9\,ms for the gate, 13.5\,ms for retrieval, 0.02\,ms for the relevance check and 1.84\,s for the verifier (Fig.~\ref{fig:latency}); a claim receives feedback a median 1.86\,s (90th percentile 2.53\,s) after its window closes.
Whisper base.en transcribes an 11\,s clip in 0.83\,s on the CPU.
The verifier uses 4.5\,GiB of GPU memory plus 0.7\,GB of host memory, and the gate and retriever 1.3--1.4\,GB.
On a board where CPU and GPU share 8\,GB, the system would need about 7.5\,GB with a headless OS: feasible, but only just.

\subsection{Model Size and Quantization}
\label{sec:size}

\begin{figure}[t]
\centering
\includegraphics[width=0.62\columnwidth]{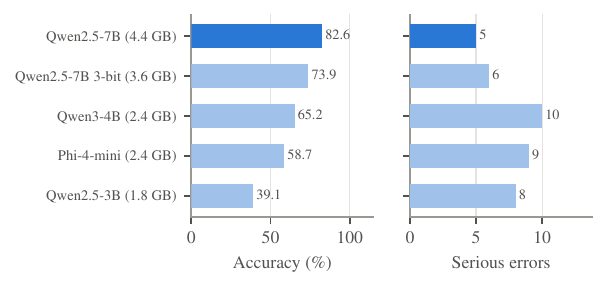}
\caption{Verifier size against accuracy and serious errors on development data (dev+dev2, 46 claims, v3 prompt).}
\label{fig:size}
\end{figure}

To fit an 8\,GB embedded board we tried smaller verifiers (Fig.~\ref{fig:size}).
None comes close: the 4B models bring back the negation error, and the 3B mostly answers that no rule applies.
The 3-bit 7B was closest on development data but makes 13 held-out serious errors.
A larger model is not a free fix either: Qwen2.5-14B (partly CPU-offloaded, 15\,s per claim) is the most accurate verifier (0.830) and fixes all five of the 7B's serious errors, but makes 13 of its own, mostly over-strict comparisons, e.g.\ ``the legal limit for DUI in Florida is point oh eight'' contradicted by a section on an ``unlawful blood-alcohol level'' of 0.08.
The v3 prompt was developed with the 7B and not retuned for the 14B.

\subsection{Second Domain: Allergy Claims}
\label{sec:allergy}

\begin{table}[t]
\centering
\caption{Allergy case study (148 test claims). * = paired 95\% interval of the difference from the full system excludes zero.}
\label{tab:allergy}
\footnotesize
\begin{tabular}{lcc}
\toprule
System & Accuracy [95\% CI] & Serious \\
\midrule
Full system (legal settings) & .878 [.824, .926] & 5 \\
Gold passage forced into top 5 & .892 & 5 \\
No retrieval (closed book) & .561* & 20* \\
BM25 retrieval & .777* & 6 \\
Hybrid retrieval & .824 & 10 \\
Top-1 / top-3 passages & .831 / .851 & 13* / 10 \\
Top-10 passages & .872 & 6 \\
Chunk context & .878 & 5 \\
300-word window (dev-tuned) & .899 & 4 \\
\bottomrule
\end{tabular}
\end{table}

To test whether the design depends on legal text, we applied the same pipeline, without retuning, to allergy statements in a doctor's visit (``if you're allergic to eggs you can't get a flu shot'').
The corpus has 280 real passages from US federal public-domain sources: CDC guidance on penicillin allergy, egg allergy, flu vaccination and vaccine contraindications; MedlinePlus pages on allergy, anaphylaxis and food allergy; the FDA allergen page; and sections of 17 FDA drug labels from openFDA.
We wrote 198 spoken-style claims (96 supported, 74 contradicted, 28 not covered) and 50 non-claim lines; every fourth claim per verdict is dev (50), the rest test (148).
The prompt keeps the v3 structure with medical wording; all other settings are the legal ones.

The full system is 87.8\% accurate with 5 serious errors (Table~\ref{tab:allergy}, Fig.~\ref{fig:domains}), and the baselines rank as in the legal domain: without retrieval the model makes 20 serious errors, mostly by calling true but recent guidance false (the 2023 CDC change for egg-allergic flu vaccination).
Two results differ.
Chunk context is as good as section context, presumably because guideline paragraphs are self-contained.
And the gate does not transfer: the traffic-trained gate fires on only 57\% of test claims, while the relevance check alone, with its threshold re-picked on dev (0.578), passes 146 of 148 test claims and 0 of 25 off-topic lines, but 10 of 25 visit-talk lines.
A domain gate is therefore the one component that must be rebuilt per domain.

\subsection{Error Analysis}
\label{sec:errors}
The five legal serious errors fall into two groups.
\emph{Right rule, wrong comparison} (3): ``the speed limit in a Texas alley is 30'' is labelled supported while quoting ``15 miles per hour in an alley''; a rear-tint limit is read as a ban; and a blue-light prohibition with an exception is read as permitting.
\emph{Wrong rule decides} (2): ``you can't arrest me over a seatbelt ticket'' is supported from \emph{Atwater}, which holds the opposite, and a Texas implied-consent claim is decided by \emph{Birchfield} because the state statute was not retrieved.
A code check comparing the numbers in \texttt{statement\_says} and \texttt{rule\_says} would catch the alley case.
Four of the five allergy errors quote the wrong sentence of the right passage, e.g.\ the ``moderate or severe illness'' precaution for ``a mild cold is not a reason to skip a vaccine''.
The live sessions showed the same pattern with units (``70 kilometers per hour'' contradicted once and supported once against a 70\,mph limit), and recognition errors from a camera microphone (``drink white driving'') were a larger source of noise than the verifier.

\section{Discussion and Limitations}
\label{sec:limits}

The results support a simple principle for high-stakes, short-form verification: \emph{give the model less to decide}.
Restricting citations to retrieved rules, deriving the verdict from a match between two restatements, and defaulting to \emph{not covered} each remove a way to be confidently wrong, at a cost mostly in coverage.
Configurations that commit to more verdicts (closed book, v2, top-10, 14B) catch more contradictions but make more serious errors (Fig.~\ref{fig:tradeoff}); a training tool might accept that trade-off, while a tool for drivers or patients should not.

The real data test the evidence base, off-topic behavior and end-to-end pipeline, but the labelled claims are synthetic and written by the system's developers; they are cleaner than real speech and may share blind spots with the design, so a labelled set of real recordings annotated by people with legal and clinical training is the most important next step.
The held-out set has 141 claims, so intervals are wide.
The corpus covers two states and 22 federal case summaries; local ordinances are out of scope, and allergy labels have not yet been reviewed by a clinician.
A \emph{supported} verdict means a retrieved rule agrees with the claim, not that the claim is correct in every circumstance.
Latency was measured on a desktop GPU, not an embedded board.

\textit{Ethics.}
\sys{} is a research prototype and its output is not legal or medical advice.
A wrong verdict could escalate a traffic stop or affect care; deployment would need expert review, clear communication that \emph{not covered} is not \emph{false}, and evaluation on speech from diverse speakers.
Running entirely on local hardware keeps sensitive audio off third-party servers.

\section{Conclusion}
\label{sec:conclusion}

We built and evaluated an on-device pipeline that checks spoken claims against rule text and stays silent when unsure.
Retrieval, a restate--quote--match prompt with the verdict mapped in code, and section-level context each significantly reduce serious errors: on held-out traffic-stop claims the system reaches 80.1\% accuracy with 3.5\% serious errors in under two seconds per claim, and on allergy claims, without retuning, 87.8\% with 3.4\%.
Next steps are a labelled real-speech test set, a domain-independent claim gate, and measurements on embedded hardware.

\bibliographystyle{IEEEtran}
\bibliography{refs}

\begin{thebibliography}{10}
\providecommand{\url}[1]{#1}
\csname url@samestyle\endcsname
\providecommand{\newblock}{\relax}
\providecommand{\bibinfo}[2]{#2}
\providecommand{\BIBentrySTDinterwordspacing}{\spaceskip=0pt\relax}
\providecommand{\BIBentryALTinterwordstretchfactor}{4}
\providecommand{\BIBentryALTinterwordspacing}{\spaceskip=\fontdimen2\font plus
\BIBentryALTinterwordstretchfactor\fontdimen3\font minus
  \fontdimen4\font\relax}
\providecommand{\BIBforeignlanguage}[2]{{%
\expandafter\ifx\csname l@#1\endcsname\relax
\typeout{** WARNING: IEEEtran.bst: No hyphenation pattern has been}%
\typeout{** loaded for the language `#1'. Using the pattern for}%
\typeout{** the default language instead.}%
\else
\language=\csname l@#1\endcsname
\fi
#2}}
\providecommand{\BIBdecl}{\relax}
\BIBdecl

\bibitem{guo2022survey}
Z.~Guo, M.~Schlichtkrull, and A.~Vlachos, ``A survey on automated
  fact-checking,'' \emph{Transactions of the Association for Computational
  Linguistics}, vol.~10, 2022.

\bibitem{nakov2021survey}
P.~Nakov, D.~Corney, M.~Hasanain, F.~Alam, T.~Elsayed,
  A.~Barr{\'o}n-Cede{\~n}o, P.~Papotti, S.~Shaar, and G.~Da~San~Martino,
  ``Automated fact-checking for assisting human fact-checkers,'' in
  \emph{Proceedings of the Thirtieth International Joint Conference on
  Artificial Intelligence (IJCAI), Survey Track}, 2021.

\bibitem{hassan2017claimbuster}
N.~Hassan, F.~Arslan, C.~Li, and M.~Tremayne, ``Toward automated fact-checking:
  Detecting check-worthy factual claims by {ClaimBuster},'' in
  \emph{Proceedings of the 23rd ACM SIGKDD International Conference on
  Knowledge Discovery and Data Mining}, 2017.

\bibitem{thorne2018fever}
J.~Thorne, A.~Vlachos, C.~Christodoulopoulos, and A.~Mittal, ``{FEVER}: a
  large-scale dataset for fact extraction and {VER}ification,'' in
  \emph{Proceedings of NAACL-HLT}, 2018.

\bibitem{chalkidis2022lexglue}
I.~Chalkidis, A.~Jana, D.~Hartung, M.~Bommarito, I.~Androutsopoulos, D.~M.
  Katz, and N.~Aletras, ``{LexGLUE}: A benchmark dataset for legal language
  understanding in {E}nglish,'' in \emph{Proceedings of the 60th Annual Meeting
  of the Association for Computational Linguistics (ACL)}, 2022.

\bibitem{guha2023legalbench}
N.~Guha, J.~Nyarko, D.~E. Ho, C.~R{\'e} \emph{et~al.}, ``{LegalBench}: A
  collaboratively built benchmark for measuring legal reasoning in large
  language models,'' in \emph{Advances in Neural Information Processing Systems
  (NeurIPS), Datasets and Benchmarks Track}, 2023.

\bibitem{dahl2024fictions}
M.~Dahl, V.~Magesh, M.~Suzgun, and D.~E. Ho, ``Large legal fictions: Profiling
  legal hallucinations in large language models,'' \emph{Journal of Legal
  Analysis}, vol.~16, 2024.

\bibitem{magesh2024hallucination}
V.~Magesh, F.~Surani, M.~Dahl, M.~Suzgun, C.~D. Manning, and D.~E. Ho,
  ``Hallucination-free? assessing the reliability of leading {AI} legal
  research tools,'' \emph{arXiv preprint arXiv:2405.20362}, 2024.

\bibitem{lewis2020rag}
P.~Lewis, E.~Perez, A.~Piktus, F.~Petroni, V.~Karpukhin, N.~Goyal,
  H.~K{\"u}ttler, M.~Lewis, W.-t. Yih, T.~Rockt{\"a}schel, S.~Riedel, and
  D.~Kiela, ``Retrieval-augmented generation for knowledge-intensive {NLP}
  tasks,'' in \emph{Advances in Neural Information Processing Systems
  (NeurIPS)}, 2020.

\bibitem{xiao2023cpack}
S.~Xiao, Z.~Liu, P.~Zhang, and N.~Muennighoff, ``{C-Pack}: Packaged resources
  to advance general {C}hinese embedding,'' \emph{arXiv preprint
  arXiv:2309.07597}, 2023.

\bibitem{robertson2009bm25}
S.~Robertson and H.~Zaragoza, ``The probabilistic relevance framework: {BM25}
  and beyond,'' \emph{Foundations and Trends in Information Retrieval}, vol.~3,
  no.~4, 2009.

\bibitem{willard2023guided}
B.~T. Willard and R.~Louf, ``Efficient guided generation for large language
  models,'' \emph{arXiv preprint arXiv:2307.09702}, 2023.

\bibitem{gerganov2023llamacpp}
G.~Gerganov \emph{et~al.}, ``llama.cpp: {LLM} inference in {C/C++},''
  \url{https://github.com/ggml-org/llama.cpp}, 2023.

\bibitem{geifman2017selective}
Y.~Geifman and R.~El-Yaniv, ``Selective classification for deep neural
  networks,'' in \emph{Advances in Neural Information Processing Systems
  (NeurIPS)}, 2017.

\bibitem{kamath2020selective}
A.~Kamath, R.~Jia, and P.~Liang, ``Selective question answering under domain
  shift,'' in \emph{Proceedings of the 58th Annual Meeting of the Association
  for Computational Linguistics (ACL)}, 2020.

\bibitem{radford2023whisper}
A.~Radford, J.~W. Kim, T.~Xu, G.~Brockman, C.~McLeavey, and I.~Sutskever,
  ``Robust speech recognition via large-scale weak supervision,'' in
  \emph{Proceedings of the 40th International Conference on Machine Learning
  (ICML)}, 2023.

\bibitem{devlin2019bert}
J.~Devlin, M.-W. Chang, K.~Lee, and K.~Toutanova, ``{BERT}: Pre-training of
  deep bidirectional transformers for language understanding,'' in
  \emph{Proceedings of NAACL-HLT}, 2019.

\bibitem{turc2019wellread}
I.~Turc, M.-W. Chang, K.~Lee, and K.~Toutanova, ``Well-read students learn
  better: On the importance of pre-training compact models,'' \emph{arXiv
  preprint arXiv:1908.08962}, 2019.

\bibitem{li2017dailydialog}
Y.~Li, H.~Su, X.~Shen, W.~Li, Z.~Cao, and S.~Niu, ``{DailyDialog}: A manually
  labelled multi-turn dialogue dataset,'' in \emph{Proceedings of the Eighth
  International Joint Conference on Natural Language Processing (IJCNLP)},
  2017.

\bibitem{qwen2024qwen25}
{Qwen Team}, ``{Qwen2.5} technical report,'' \emph{arXiv preprint
  arXiv:2412.15115}, 2024.

\bibitem{koehn2004significance}
P.~Koehn, ``Statistical significance tests for machine translation
  evaluation,'' in \emph{Proceedings of EMNLP}, 2004.

\bibitem{dror2018hitchhiker}
R.~Dror, G.~Baumer, S.~Shlomov, and R.~Reichart, ``The hitchhiker's guide to
  testing statistical significance in natural language processing,'' in
  \emph{Proceedings of the 56th Annual Meeting of the Association for
  Computational Linguistics (ACL)}, 2018.

\bibitem{wilson1927}
E.~B. Wilson, ``Probable inference, the law of succession, and statistical
  inference,'' \emph{Journal of the American Statistical Association}, vol.~22,
  no. 158, 1927.

\end{thebibliography}

\end{document}